\pdfoutput=1

\documentclass[11pt]{article}

\usepackage[]{acl}

\usepackage{times}
\usepackage{latexsym}

\usepackage[T1]{fontenc}

\usepackage[utf8]{inputenc}

\usepackage{microtype}

\usepackage{graphicx}

\title{ArgLegalSumm: Improving Abstractive Summarization of Legal Documents with Argument Mining}

\author{Mohamed Elaraby, Diane Litman \\
        University of Pittsburgh \\ Pittsburgh, PA, USA \\ \texttt{\{mse30,dlitman\}@pitt.edu}}

\begin{document}
\maketitle
\begin{abstract}
A challenging task when generating summaries
of legal documents is the ability to address their
argumentative nature. We introduce a simple
technique to capture the argumentative structure of legal documents by integrating \textit{argument role labeling} into the summarization process. Experiments with pretrained language models show that our proposed approach improves performance over strong baselines.
\end{abstract}

\section{Introduction}

Abstractive summarization has made great progress by leveraging large 
pretrained language models such as BART \cite{lewis2020bart}, T5 \cite{raffel2020exploring}, Pegasus \cite{zhang2020pegasus}, and Longformer \cite{beltagy2020longformer}. These models leverage large scale datasets such as CNN-DailyMail \cite{hermann2015teaching}, PubMed \cite{cohan2018discourse}, and New York Times \cite{sandhaus2008new}.   Unlike news 
and scientific texts, which contain specific formatting such as topic sentences and abstracts,  legal cases 
contain implicit argument structure spreading across long  
texts  \cite{xu2021toward}. Current abstractive summarization models do not take into account the argumentative structure of the text, which poses a challenge towards effective abstractive summarization of legal documents. 



In this work, we bridge the gap between prior research focusing on summarizing legal documents through extracting argument roles of  legal text \cite{grover2003automatic, xu2021toward, saravanan2010identification}, and prior research focused on producing abstractive summaries of legal text \cite{feijo2019summarizing, bajaj2021long}. Our work proposes a  technique that {\it blends argument role mining  and abstractive summarization}, which hasn't been explored extensively in the literature. 

\begin{figure}[h]
\begin{center}

 \includegraphics[width=7.4cm, height=3.4cm]{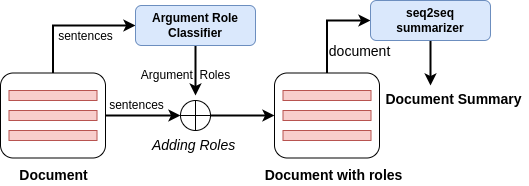}
 \caption{\label{sys_diag} Overview of our approach.}
 \end{center}
\end{figure}

Figure \ref{sys_diag} describes the main flow of our approach, which decomposes the summarization process into two tasks.
 First, each sentence in the document is assigned an argument role by using an independent  model. Then, the predicted roles are blended with the original document's sentences and fed into a sequence to sequence-based abstractive summarizer. 

Our contributions are as follows:
\textbf{(a)} We propose a simple 
technique to create an  argument-aware neural abstractive summarizer. 
\textbf{(b)} We show the effectiveness of this technique in improving  legal document summarization. 
\textbf{(c)} We make our code \footnote{\url{https://github.com/EngSalem/arglegalsumm/}} and  argument role  annotations 
freely available\footnote{The data was obtained through an agreement with the Canadian Legal Information Institute (CanLII) (\url{https://www.canlii.org/en/})}.  

\section{Related Work}

\textbf{Legal Document Summarization.}
Prior research 
has mainly focused on extractive techniques \cite{galgani2015summarization, anand2019effective,jain2021automatic}, 
exploiting features such as the document structure and prior knowledge of legal terms to extract salient sentences that represent the summary of the legal text. Recent research 
has also shifted gears to abstractive techniques due to their superiority to extractive methods on automatic measures such as ROUGE \cite{feijo2019summarizing}. These abstractive techniques benefited from  neural  
models such as pointer generator networks \cite{see2017get} 
(utilized in legal public opinion summarization \cite{huang2020legal}) and 
transformer-based sequence to sequence encoder-decoder architectures such as BART \cite{lewis2020bart} and Longformer \cite{beltagy2020longformer} (employed to summarize long legal documents \cite{moro2022semantic}). However, the current abstractive approaches 
ignore the argumentative structure of the legal text. In our work, we 
combine both the rich argumentative structure of legal documents and state-of-the-art abstractive summarization models. 

\textbf{Argument Mining.}
Argument mining aims to represent the argumentative structure of a text in a graph structure that contains the argument roles and their relationship to each other. Constructing the graphs usually involves several steps: extracting argument units, classifying the argument roles of the units, and detecting the relationship between different argument roles.  Recently, contextualized embeddings were employed to improve argument role labeling \cite{reimers2019classification, elaraby2021self}. In many domains, argument roles are classified into 
 \textit{claims}, \textit{major claims}, and \textit{premises} as proposed in \citet{stab2014identifying}. Alternatively,  \citet{xu2021toward} proposed  to classify the argument roles in legal documents to \textit{Issues}, \textit{Reasons}, and \textit{Conclusions} which fits the legal text structure. 
We use the same set  of legal argument role labels in our work, and use contextualized embeddings to automatically predict them.

\textbf{Argument Mining and Summarization.}
Prior research integrating argument mining and summarization has mainly focused on extracting chunks of text that contain \textit{key argument units} \cite{barkers, bar2020quantitative, friedman2021overview}. Combining argument mining and abstractive summarization has received less attention in the literature. Recently, \citet{fabbri2021convosumm} proposed a dialogue summarization dataset with argument information. In their work, the authors included argument information in abstractive summarization by linearizing the argument graph to a textual format, which is used to train an \textit{encoder-decoder} summarization model. However, their proposed approach didn't  improve over  \textit{encoder-decoder} baselines. 
We propose an alternative method that relies on argument roles only, which shows higher improvements over \textit{encoder-decoder} baselines. 

\section{Dataset and Methods}
\subsection{Dataset \footnote{See Appendix \ref{sec:appex_stats} for more detailed statistics.}}

\label{sec:Dataset}
{\bf Texts.} 
Our dataset is composed of $1049$ 
legal cases and summary pairs, obtained through an agreement with the Canadian Legal Information Institute. 
We split these pairs into training ($839$ pairs, about  $80\%$),
validation ($106$ pairs, about $10\%$)
and testing ($104$ pairs, about $10\%$) datasets.

\textbf{Document 
Lengths.} 
The maximum length of our input documents is $26k$ words, which motivates us to include \textit{encoder-decoder} architectures like Longformer that can encode long documents. 

\begin{figure}[t]
\begin{center}
\includegraphics[width=8cm, height=5cm]{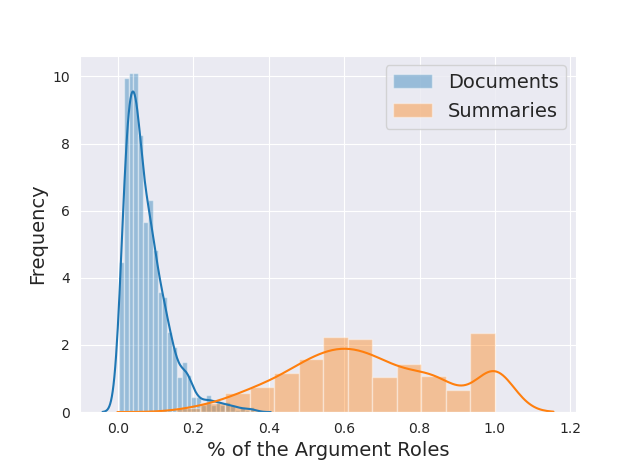}
\caption{\label{ircs_dist} Argumentative sentence \% in dataset.}
\end{center}
\end{figure}

\textbf{Argument Role Annotations.} 
The dataset was annotated prior to our study, using the IRC taxonomy of three legal argument roles described in \citet{xu2021toward}:  
\textbf{Issues} (legal questions which a court addressed in the document), \textbf{Reasons} (pieces of text which indicate why the court reached the specific conclusions) , and  \textbf{Conclusions} (court’s decisions for the corresponding issues). Figure \ref{ircs_dist} shows the distribution of the percentage of sentences annotated with an argumentative role across the articles and reference human summaries. 
We can see that while only a small percent of the sentences in the original articles are annotated as argument units, argumentative units dominate the reference summaries. Thus, we hypothesize that augmenting the summarization model with argument roles in the input text should  improve the generated summaries.

\subsection{Methods}
\label{sec:method}
\textbf{Special Tokens Approach.} We designate special marker tokens to distinguish between different argument roles.  In prior 
research, \citet{deyoung2021ms2} used markers such as \texttt{<evidence>,</evidence>} to highlight evidence sentences in summarizing medical scientific documents, while \citet{khalifa2021bag} used \texttt{<neg>,</neg>} to mark negation phrases in dialogue summarization. However, we explore the impact of changing token granularity by experimenting with two sets of special  tokens. First, we introduce \texttt{<IRC>,</IRC>} to distinguish between argumentative and non-argumentative sentences. 
Second, we broaden the list of the proposed special tokens  to differentiate between the three argument roles mentioned in Section \ref{sec:Dataset}. We assign each argument role two unique tokens to highlight its boundaries in the text, e.g., we use 
\texttt{<Reason>,</Reason>} to highlight the \textbf{reason} roles.
Table \ref{tab:examples} shows examples using tokens   to highlight an argumentative  sentence (top) versus a  specific argumentative role (bottom). 

\begin{table}[t]

\begin{tabular}{|p{0.45\textwidth}|}
\hline
Example                                                                                                                                                                                        \\ \hline
\textbf{\textless{}IRC\textgreater} He also found “on the strong balance of probabilities,” that the late Mrs. Scott intended to make an inter vivos gift to Ms. Akerley. \textbf{\textless{}/IRC\textgreater{}} \\ \hline
\textbf{\textless{}Issue\textgreater} {[}6{]} Mr. Comeau appeals, arguing that the probate court judge erred: \textbf{\textless{}/Issue\textgreater{}}                                                          \\ \hline
\end{tabular}
\caption{\label{tab:examples}  Different special tokens  for  argument roles.}
\end{table}

\begin{figure}[t]
\begin{center}

 \includegraphics[width=7.5cm, height=6.25cm]{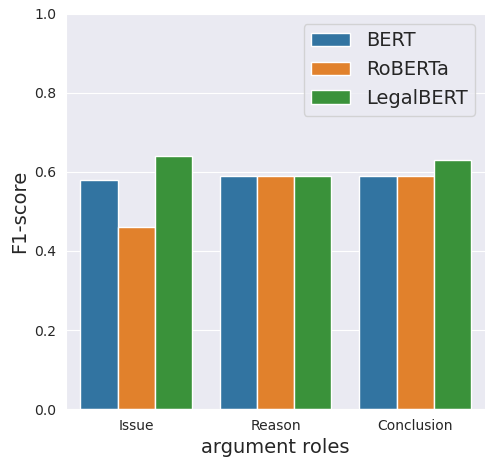}
 \end{center}
 \caption{\label{classification_fig} Argument role detection results. }
\end{figure}

\textbf{Sentence-level Argument Role Mining.}  Our data's argument role annotation  is at the sentence level, thus, we perform 
sentence-level classification experiments using the same data splits employed in summarization to detect argument roles.\footnote{See Appendix \ref{sec:hyper} for argument mining training details.}
We experiment with several contextualized embedding-based techniques, namely \textit{BERT} \cite{devlin2019bert}, \textit{RoBERTa} \cite{liu2019roberta}, and \textit{legalBERT} \cite{zhengguha2021}.  We employ these models to predict sentences' 
argument roles (issues, reasons, conclusions, or non-argumentative). Figure \ref{classification_fig} shows that \textit{legalBERT} achieved the best classification results. We achieved a macro average F1 of $63.4\%$ in argument role classification and $71\%$ in binary classification using \textit{legalBERT}. 
Thus, we rely on its 
predictions to integrate argument roles into summarization below.

\section{Experiments and Results}

Our experiments are conducted in two settings: assuming argument roles are manually labeled (which we  refer to as {\it oracle}) versus predicting  argument role labels (referred to as {\it predicted}). 

\subsection{Baselines}

We compare our proposed argument-aware summarization method to two sets of baselines\footnote{See Appendix \ref{sec:hyper} for summarization training details.}:

 \textbf{Extractive Baseline.} We employ the unsupervised method of \citet{miller2019leveraging}. The model leverages BERT embeddings and k-means to extract salient sentences based on their proximity to cluster centroids.

\textbf{Abstractive Baselines.} 
\textbf{Vanilla BART-Large} refers to finetuning BART-large on our dataset.  For \textbf{Vanilla LED-base}, similarly to BART, the model is finetuned using Longformer-base checkpoint.

 \makeatletter
\def\hlinewd#1{%
\noalign{\ifnum0=`}\fi\hrule \@height #1 \futurelet
\reserved@a\@xhline}
\makeatother

\begin{table*}[]
\begin{center}
    
\begin{tabular}{p{0.11\textwidth}p{0.15\textwidth}lccc}
\hlinewd{1.5pt}
     \textbf{Setting} &                                 \textbf{Experiment}    & \textbf{Model}                   & \multicolumn{1}{l}{\textbf{Rouge-1}} & \multicolumn{1}{l}{\textbf{Rouge-2}} & \multicolumn{1}{l}{\textbf{Rouge-L}} \\ \hlinewd{1.5pt}
                                          & & Unsupervised Extractive BERT     & 37.71                                 & 14.77                                 & {\color[HTML]{212121} 36.41}           \\  
                                         & & Vanilla BART-Large               & 47.93                                 & 22.34                                 & 44.74                                 \\ 
& \textbf{Baselines}      & Vanilla LED-base                 & 49.56                                 & 22.75                                 & 46.48                                 \\  \hline
                                          &\textbf{arg-BART-Large} & BART-Large + 2 markers  & 47.11                                 & 21.77                                 & 43.12                                 \\ 
\textbf{Oracle} &   &BART-Large +  6 markers  & 46.80                                  & 22.14                                 & 44.14                   \\  \cline{2-6}
                                       &   & LED-base + 2 markers    & 49.64                                 & \textbf{26.81}                        & 46.70                                  \\ 
                                         & \textbf{arg-LED-base} & LED-base + 6 markers    & \textbf{50.53}                        & 26.31                             & \textbf{47.90}    \\ \hline
                                     &    & LED-base + 2 markers  & 49.50                                  & {\it 26.46}                                 & 46.60                                  \\ 
\textbf{Predicted} & \textbf{arg-LED-base}  & LED-base + 6 markers  & {\it 50.23}                                 & 26.29                                 & {\it 47.49}                                 \\ \hlinewd{1.5pt}
\end{tabular}
\end{center}
\caption{\label{res} Summarization results 
on the test set. Best results {\bf bolded}. Best results using predicted  roles {\it italicized}.}
\end{table*}

\begin{table*}[t]

\begin{center}
\begin{tabular}{cllll}

\hlinewd{1.5pt}
{\bf Model}                 & \multicolumn{1}{l}{{\bf Rouge-1}} & \multicolumn{1}{l}{{\bf Rouge-2}} & \multicolumn{1}{l}{{\bf Rouge-L}} & {{\bf Mean Summary Length}} \\ \hlinewd{1.5pt}
Vanilla LED-base      & 48.25                        & 21.60                        & 44.88    &\quad \quad \quad \quad 267                    \\ 
arg-LED-base + 2  markers & 50.43                        & \textbf{27.76}               & 47.05     &\quad \quad \quad \quad 156                 \\ 
arg-LED-base + 6 markers  & \textbf{50.73}               & 27.29                        & \textbf{47.30}     & \quad \quad \quad \quad 174         \\ \hlinewd{1.5pt}
\end{tabular}

\caption{\label{eval_arg} Comparing  Longformer (LED) summaries with sentences labeled as argumentative in reference summary.}
\end{center}
\end{table*}

\subsection{Results and Discussion}

Table \ref{res} evaluates the results of the different summarization models using Rouge-1, Rouge-2, and Rouge-L scores.\footnote{See Appendix \ref{sec:appendix} for example generated summaries.} We refer to BART and Longformer augmented with argument roles as \textbf{arg-BART-Large} and \textbf{arg-LED-base}, respectively.  We use \textit{2 markers} to denote the use of binary special tokens (i.e; \texttt{<IRC>,</IRC>}) and \textit{6 markers} to refer to the full set of argument role  tokens. 
We include  two 
markers sets
to examine whether it's necessary to include explicit argument roles or if it's sufficient to highlight argumentative text only. 

\begin{figure}[t]
\begin{center}
\includegraphics[width=8.25cm, height=6.5cm]{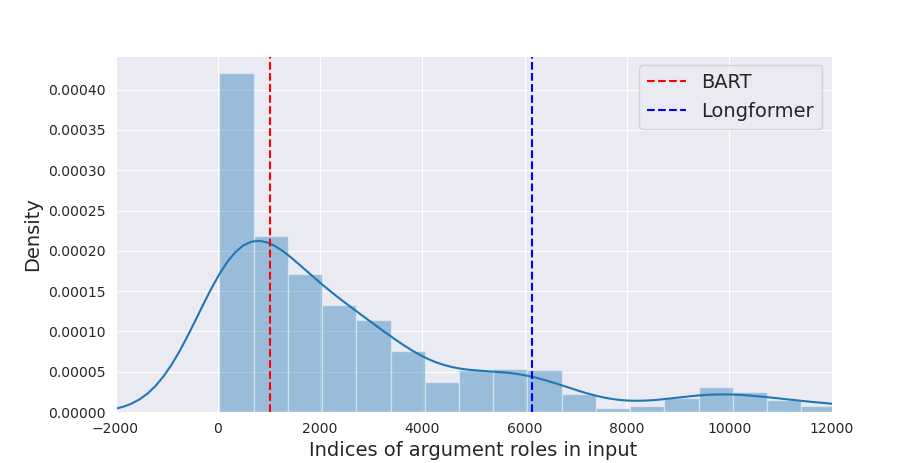}
\caption{\label{ircs_ixs} Argument position distribution in the input.}
\end{center}
\end{figure}

We first evaluate the models augmented with the {\it manually labeled argument roles} to examine whether adding argument information has the potential to improve over the baselines. The {\it oracle} results 
in Table \ref{res} show that  \textbf{arg-LED-base} improves 
performance in terms of Rouge-1, Rouge-2, and Rouge-L \cite{lin2004rouge} by approximately $1$, $4$, and $1.5$ points, respectively, over the vanilla LED-base baseline when using the \textit{6
markers}. The \textit{2 markers} set showed marginal improvements on Rouge-1 and Rouge-L, but showed  $4$ Rouge-2 points improvement over the  baseline. These results indicate that representing  argument roles using  fine-grained labels is the most effective in improving LED model output.  In contrast,  \textbf{arg-BART-Large}  didn't show improvements over the vanilla BART-Large baseline. We hypothesize that this is due to the  sparsity of the argumentative sentences in the input documents 
(recall Figure \ref{ircs_dist}). Since Longformer  can encode more words, it was likely able to capture more argument markers added to the input, 
increasing the model's ability to grasp the argument structure of the legal text. To validate this hypothesis, we analyze the positions of each argument role across the input articles. Figure \ref{ircs_ixs} shows that the argument roles are distributed across the article and not centered around a unique position. This aligns with our hypothesis that the Longformer's encoding limit \textit{(blue dashed line)} can cover significantly more roles when compared to the BART's encoding limit \textit{(red dashed line)}.

Next, we evaluate the summarization using {\it predicted argument roles} obtained from our 
classifier (Section \ref{sec:method}). We evaluate the Longformer summarization model only, since BART didn't show oracle improvements. The last two rows of Table \ref{res} (the {\it predicted} results) show that including predicted argument roles  showed consistent improvements with the manually labeled ones (oracle). 
The results showed a minimal drop in Rouge scores ranging from  
$0.02-0.41$ points when using the predicted argument roles both in the \textit{6 markers} and \textit{2 markers} cases,  which indicates the effectiveness of our approach in practical scenarios.

Finally, to  estimate the argumentativeness of the  LED-based (oracle) summaries, we evaluate them  against 
a summary containing only the sentences manually annotated as an IRC sentence  
in the 
reference summary.

Table \ref{eval_arg} shows that adding argument role markers increases the overlap between the generated summaries and the argumentative sentence subset from  the reference summaries, suggesting that our proposed model's gains are mainly obtained from an increase in argumentativeness of the generated summaries. The generated summaries from our \textbf{arg-LED-base} are  lower in length compared to the baseline. We hypothesize that this is due to the focus on argument roles mainly, discarding some of the non-argumentative content.  \footnote{See Appendix \ref{sec:appendix} for an illustrative example.}

\section{Conclusion and Future Work}
We proposed to utilize argument roles in the abstractive summarization of legal documents to accommodate their  argumentative structure. Our experiments with state-of-the-art encoder-decoder models showed that including argument role information 
can improve the ROUGE scores of summarization models capable of handling long documents. Specifically, improved results were achieved using  Longformer with input documents augmented with argument roles (highlighted using special marker tokens), and this finding was  robust across two special token schemes. We also showed that using predicted argument roles 
showed consistent improvements to using the manually labeled ones.
In future work, we plan to explore methods to unify argument mining and summarization to reduce the computational resources necessary to host two 
models. 

\section*{Acknowledgements}
This material is based upon work supported by the National Science Foundation under Grant No. 2040490 and by Amazon. We would like to thank the members of both the Pitt AI Fairness and Law Project and the Pitt PETAL group, as well as the anonymous reviewers, for valuable comments in improving this work.

\bibliography{anthology,custom}

\begin{thebibliography}{33}
\expandafter\ifx\csname natexlab\endcsname\relax\def\natexlab#1{#1}\fi

\bibitem[{Anand and Wagh(2019)}]{anand2019effective}
Deepa Anand and Rupali Wagh. 2019.
\newblock Effective deep learning approaches for summarization of legal texts.
\newblock \emph{Journal of King Saud University-Computer and Information
  Sciences}.

\bibitem[{Bajaj et~al.(2021)Bajaj, Dangati, Krishna, Kumar, Uppaal, Windsor,
  Brenner, Dotterrer, Das, and McCallum}]{bajaj2021long}
Ahsaas Bajaj, Pavitra Dangati, Kalpesh Krishna, Pradhiksha~Ashok Kumar, Rheeya
  Uppaal, Bradford Windsor, Eliot Brenner, Dominic Dotterrer, Rajarshi Das, and
  Andrew McCallum. 2021.
\newblock Long document summarization in a low resource setting using
  pretrained language models.
\newblock In \emph{Proceedings of the 59th Annual Meeting of the Association
  for Computational Linguistics and the 11th International Joint Conference on
  Natural Language Processing: Student Research Workshop}, pages 71--80.

\bibitem[{Bar-Haim et~al.(2020)Bar-Haim, Kantor, Eden, Friedman, Lahav, and
  Slonim}]{bar2020quantitative}
Roy Bar-Haim, Yoav Kantor, Lilach Eden, Roni Friedman, Dan Lahav, and Noam
  Slonim. 2020.
\newblock Quantitative argument summarization and beyond: Cross-domain key
  point analysis.
\newblock In \emph{Proceedings of the 2020 Conference on Empirical Methods in
  Natural Language Processing (EMNLP)}, pages 39--49.

\bibitem[{Barker et~al.()Barker, Paramita, Funk, Kurtic, Aker, Foster, Hepple,
  and Gaizauskas}]{barkers}
Emma Barker, Monica Paramita, Adam Funk, Emina Kurtic, Ahmet Aker, Jonathan
  Foster, Mark Hepple, and Robert Gaizauskas.
\newblock What’s the issue here?: Task-based evaluation of reader comment
  summarization systems.

\bibitem[{Beltagy et~al.(2020)Beltagy, Peters, and
  Cohan}]{beltagy2020longformer}
Iz~Beltagy, Matthew~E Peters, and Arman Cohan. 2020.
\newblock Longformer: The long-document transformer.
\newblock \emph{arXiv preprint arXiv:2004.05150}.

\bibitem[{Cohan et~al.(2018)Cohan, Dernoncourt, Kim, Bui, Kim, Chang, and
  Goharian}]{cohan2018discourse}
Arman Cohan, Franck Dernoncourt, Doo~Soon Kim, Trung Bui, Seokhwan Kim, Walter
  Chang, and Nazli Goharian. 2018.
\newblock A discourse-aware attention model for abstractive summarization of
  long documents.
\newblock In \emph{Proceedings of the 2018 Conference of the North American
  Chapter of the Association for Computational Linguistics: Human Language
  Technologies, Volume 2 (Short Papers)}, pages 615--621.

\bibitem[{Devlin et~al.(2019)Devlin, Chang, Lee, and
  Toutanova}]{devlin2019bert}
Jacob Devlin, Ming-Wei Chang, Kenton Lee, and Kristina Toutanova. 2019.
\newblock Bert: Pre-training of deep bidirectional transformers for language
  understanding.
\newblock In \emph{Proceedings of the 2019 Conference of the North American
  Chapter of the Association for Computational Linguistics: Human Language
  Technologies, Volume 1 (Long and Short Papers)}, pages 4171--4186.

\bibitem[{DeYoung et~al.(2021)DeYoung, Beltagy, van Zuylen, Kuehl, and
  Wang}]{deyoung2021ms2}
Jay DeYoung, Iz~Beltagy, Madeleine van Zuylen, Bailey Kuehl, and Lucy Wang.
  2021.
\newblock Msˆ2: Multi-document summarization of medical studies.
\newblock In \emph{Proceedings of the 2021 Conference on Empirical Methods in
  Natural Language Processing}, pages 7494--7513.

\bibitem[{Elaraby and Litman(2021)}]{elaraby2021self}
Mohamed Elaraby and Diane Litman. 2021.
\newblock Self-trained pretrained language models for evidence detection.
\newblock In \emph{Proceedings of the 8th Workshop on Argument Mining}, pages
  142--147.

\bibitem[{Fabbri et~al.(2021)Fabbri, Rahman, Rizvi, Wang, Li, Mehdad, and
  Radev}]{fabbri2021convosumm}
Alexander~Richard Fabbri, Faiaz Rahman, Imad Rizvi, Borui Wang, Haoran Li,
  Yashar Mehdad, and Dragomir Radev. 2021.
\newblock Convosumm: Conversation summarization benchmark and improved
  abstractive summarization with argument mining.
\newblock In \emph{Proceedings of the 59th Annual Meeting of the Association
  for Computational Linguistics and the 11th International Joint Conference on
  Natural Language Processing (Volume 1: Long Papers)}, pages 6866--6880.

\bibitem[{Feijo and Moreira(2019)}]{feijo2019summarizing}
Diego Feijo and Viviane Moreira. 2019.
\newblock Summarizing legal rulings: Comparative experiments.
\newblock In \emph{proceedings of the international conference on recent
  advances in natural language processing (RANLP 2019)}, pages 313--322.

\bibitem[{Friedman et~al.(2021)Friedman, Dankin, Hou, Aharonov, Katz, and
  Slonim}]{friedman2021overview}
Roni Friedman, Lena Dankin, Yufang Hou, Ranit Aharonov, Yoav Katz, and Noam
  Slonim. 2021.
\newblock Overview of the 2021 key point analysis shared task.
\newblock In \emph{Proceedings of the 8th Workshop on Argument Mining}, pages
  154--164.

\bibitem[{Galgani et~al.(2015)Galgani, Compton, and
  Hoffmann}]{galgani2015summarization}
Filippo Galgani, Paul Compton, and Achim Hoffmann. 2015.
\newblock Summarization based on bi-directional citation analysis.
\newblock \emph{Information processing \& management}, 51(1):1--24.

\bibitem[{Grover et~al.(2003)Grover, Hachey, Hughson, and
  Korycinski}]{grover2003automatic}
Claire Grover, Ben Hachey, Ian Hughson, and Chris Korycinski. 2003.
\newblock Automatic summarisation of legal documents.
\newblock In \emph{Proceedings of the 9th international conference on
  Artificial intelligence and law}, pages 243--251.

\bibitem[{Hermann et~al.(2015)Hermann, Kocisky, Grefenstette, Espeholt, Kay,
  Suleyman, and Blunsom}]{hermann2015teaching}
Karl~Moritz Hermann, Tomas Kocisky, Edward Grefenstette, Lasse Espeholt, Will
  Kay, Mustafa Suleyman, and Phil Blunsom. 2015.
\newblock Teaching machines to read and comprehend.
\newblock \emph{Advances in neural information processing systems},
  28:1693--1701.

\bibitem[{Huang et~al.(2020)Huang, Yu, Guo, Yu, and Xian}]{huang2020legal}
Yuxin Huang, Zhengtao Yu, Junjun Guo, Zhiqiang Yu, and Yantuan Xian. 2020.
\newblock Legal public opinion news abstractive summarization by incorporating
  topic information.
\newblock \emph{International Journal of Machine Learning and Cybernetics},
  11(9):2039--2050.

\bibitem[{Jain et~al.(2021)Jain, Borah, and Biswas}]{jain2021automatic}
Deepali Jain, Malaya~Dutta Borah, and Anupam Biswas. 2021.
\newblock Automatic summarization of legal bills: A comparative analysis of
  classical extractive approaches.
\newblock In \emph{2021 International Conference on Computing, Communication,
  and Intelligent Systems (ICCCIS)}, pages 394--400. IEEE.

\bibitem[{Khalifa et~al.(2021)Khalifa, Ballesteros, and
  Mckeown}]{khalifa2021bag}
Muhammad Khalifa, Miguel Ballesteros, and Kathleen Mckeown. 2021.
\newblock A bag of tricks for dialogue summarization.
\newblock In \emph{Proceedings of the 2021 Conference on Empirical Methods in
  Natural Language Processing}, pages 8014--8022.

\bibitem[{Lewis et~al.(2020)Lewis, Liu, Goyal, Ghazvininejad, Mohamed, Levy,
  Stoyanov, and Zettlemoyer}]{lewis2020bart}
Mike Lewis, Yinhan Liu, Naman Goyal, Marjan Ghazvininejad, Abdelrahman Mohamed,
  Omer Levy, Veselin Stoyanov, and Luke Zettlemoyer. 2020.
\newblock Bart: Denoising sequence-to-sequence pre-training for natural
  language generation, translation, and comprehension.
\newblock In \emph{Proceedings of the 58th Annual Meeting of the Association
  for Computational Linguistics}, pages 7871--7880.

\bibitem[{Lin(2004)}]{lin2004rouge}
Chin-Yew Lin. 2004.
\newblock Rouge: A package for automatic evaluation of summaries.
\newblock In \emph{Text summarization branches out}, pages 74--81.

\bibitem[{Liu et~al.(2019)Liu, Ott, Goyal, Du, Joshi, Chen, Levy, Lewis,
  Zettlemoyer, and Stoyanov}]{liu2019roberta}
Yinhan Liu, Myle Ott, Naman Goyal, Jingfei Du, Mandar Joshi, Danqi Chen, Omer
  Levy, Mike Lewis, Luke Zettlemoyer, and Veselin Stoyanov. 2019.
\newblock Roberta: A robustly optimized bert pretraining approach.
\newblock \emph{arXiv preprint arXiv:1907.11692}.

\bibitem[{Miller(2019)}]{miller2019leveraging}
Derek Miller. 2019.
\newblock Leveraging bert for extractive text summarization on lectures.
\newblock \emph{arXiv preprint arXiv:1906.04165}.

\bibitem[{Moro and Ragazzi(2022)}]{moro2022semantic}
Gianluca Moro and Luca Ragazzi. 2022.
\newblock Semantic self-segmentation for abstractive summarization of long
  legal documents in low-resource regimes.
\newblock In \emph{Proceedings of the Thirty-Six AAAI Conference on Artificial
  Intelligence, Virtual}, volume~22.

\bibitem[{Raffel et~al.(2020)Raffel, Shazeer, Roberts, Lee, Narang, Matena,
  Zhou, Li, and Liu}]{raffel2020exploring}
Colin Raffel, Noam Shazeer, Adam Roberts, Katherine Lee, Sharan Narang, Michael
  Matena, Yanqi Zhou, Wei Li, and Peter~J Liu. 2020.
\newblock Exploring the limits of transfer learning with a unified text-to-text
  transformer.
\newblock \emph{Journal of Machine Learning Research}, 21:1--67.

\bibitem[{Reimers et~al.(2019)Reimers, Schiller, Beck, Daxenberger, Stab, and
  Gurevych}]{reimers2019classification}
Nils Reimers, Benjamin Schiller, Tilman Beck, Johannes Daxenberger, Christian
  Stab, and Iryna Gurevych. 2019.
\newblock Classification and clustering of arguments with contextualized word
  embeddings.
\newblock In \emph{Proceedings of the 57th Annual Meeting of the Association
  for Computational Linguistics}, pages 567--578.

\bibitem[{Sandhaus(2008)}]{sandhaus2008new}
Evan Sandhaus. 2008.
\newblock The new york times annotated corpus.
\newblock \emph{Linguistic Data Consortium, Philadelphia}, 6(12):e26752.

\bibitem[{Saravanan and Ravindran(2010)}]{saravanan2010identification}
M~Saravanan and Balaraman Ravindran. 2010.
\newblock Identification of rhetorical roles for segmentation and summarization
  of a legal judgment.
\newblock \emph{Artificial Intelligence and Law}, 18(1):45--76.

\bibitem[{See et~al.(2017)See, Liu, and Manning}]{see2017get}
Abigail See, Peter~J Liu, and Christopher~D Manning. 2017.
\newblock Get to the point: Summarization with pointer-generator networks.
\newblock In \emph{Proceedings of the 55th Annual Meeting of the Association
  for Computational Linguistics (Volume 1: Long Papers)}, pages 1073--1083.

\bibitem[{Stab and Gurevych(2014)}]{stab2014identifying}
Christian Stab and Iryna Gurevych. 2014.
\newblock Identifying argumentative discourse structures in persuasive essays.
\newblock In \emph{Proceedings of the 2014 Conference on Empirical Methods in
  Natural Language Processing (EMNLP)}, pages 46--56.

\bibitem[{Wolf et~al.(2019)Wolf, Debut, Sanh, Chaumond, Delangue, Moi, Cistac,
  Rault, Louf, Funtowicz et~al.}]{wolf2019huggingface}
Thomas Wolf, Lysandre Debut, Victor Sanh, Julien Chaumond, Clement Delangue,
  Anthony Moi, Pierric Cistac, Tim Rault, R{\'e}mi Louf, Morgan Funtowicz,
  et~al. 2019.
\newblock Huggingface's transformers: State-of-the-art natural language
  processing.
\newblock \emph{arXiv preprint arXiv:1910.03771}.

\bibitem[{Xu et~al.(2021)Xu, Savelka, and Ashley}]{xu2021toward}
Huihui Xu, Jaromir Savelka, and Kevin~D Ashley. 2021.
\newblock Toward summarizing case decisions via extracting argument issues,
  reasons, and conclusions.
\newblock In \emph{Proceedings of the eighteenth international conference on
  artificial intelligence and law}, pages 250--254.

\bibitem[{Zhang et~al.(2020)Zhang, Zhao, Saleh, and Liu}]{zhang2020pegasus}
Jingqing Zhang, Yao Zhao, Mohammad Saleh, and Peter Liu. 2020.
\newblock Pegasus: Pre-training with extracted gap-sentences for abstractive
  summarization.
\newblock In \emph{International Conference on Machine Learning}, pages
  11328--11339. PMLR.

\bibitem[{Zheng et~al.(2021)Zheng, Guha, Anderson, Henderson, and
  Ho}]{zhengguha2021}
Lucia Zheng, Neel Guha, Brandon~R. Anderson, Peter Henderson, and Daniel~E. Ho.
  2021.
\newblock \href {http://arxiv.org/abs/2104.08671} {When does pretraining help?
  assessing self-supervised learning for law and the casehold dataset}.
\newblock In \emph{Proceedings of the 18th International Conference on
  Artificial Intelligence and Law}. Association for Computing Machinery.

\end{thebibliography}
\bibliographystyle{acl_natbib}

\appendix

\section{Data statistics}
\label{sec:appex_stats}

\subsection{Length statistics}
\label{sec:length}

\begin{figure}[ht]
\begin{center}

 \includegraphics[width=7.25cm, height=6cm]{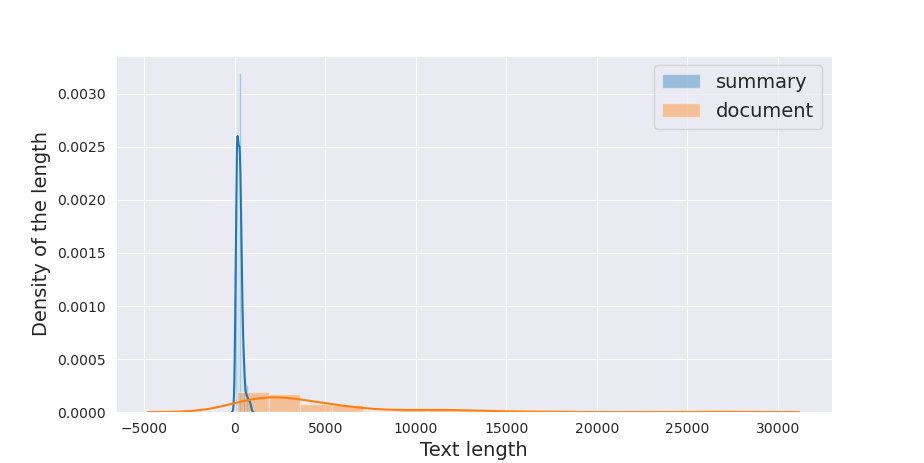}
 \end{center}
 \caption{\label{article_summ_len} Distribution of article and summary length. }
\end{figure}

Figure \ref{article_summ_len} shows the distribution of document and summary lengths. The summaries' lengths are centered around a mean of $255$ words, with a maximum length of $850$ words. The $90th$ percentile of summary length is $~490$ words. Thus we chose the maximum length of generated summary in our hyperparameters to be set to $512$ words. 
Unlike the summaries, the documents are more spread across the distribution. In our analysis, we found that the mean document length is $4180$ words, while the maximum document length is $26235$ words. 

\subsection{Argument role distribution}
\label{sec:argm_roles}
\begin{figure}[ht]
\begin{center}

\includegraphics[width=8cm, height=6.5cm]{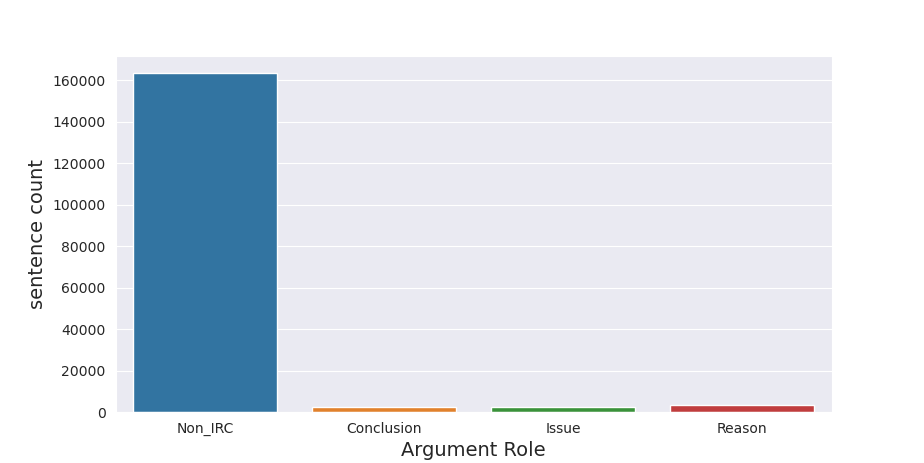}
\end{center}
\caption{\label{role_count} Count of argument roles across training set. \textit{non\_IRC} refers to non-argumentative sentences.}
\end{figure}

While they are essential to legal cases, argument roles represent a small percentage of the document.  Figure \ref{role_count} shows the high imbalance of the manually annotated argumentative versus non-argumentative sentences in our training set, which poses a challenge in building a sentence level classifier of argument roles. In our analysis we found that the non-argumentative sentences count is approximately $1000\times$  the argumentative sentences, which we use to adjust class weights in our learning objective. 

\section{Training details and hyperparameters}
\label{sec:hyper}

All experiments use the model implementations provided in the \textit{Huggingface library} \cite{wolf2019huggingface}. We initialize all our models with the same learning rate of $2e^{-5}$. We train both our summarization and argument role classification models for $10$ epochs with early stopping with $3$ epoch patience. 
For training summarization models, we set the maximum summary length to $512$ words. We truncate the input length to $1024$ words for the \textit{BART} model while truncating the input length to $6144$ words for the \textit{Longformer} due to our GPU limitation \footnote{We use Quadro RTX 5000  which has 16 GB RAM.}. We pick our best model based on its \textit{ROUGE-2} \cite{lin2004rouge} score on the validation set.  For the classification models introduced in Section~\ref{sec:method}, due to the high imbalance of our argumentative labels, we introduce fixed class weights to our cross-entropy loss. For argumentative sentences, we modify the \textit{cross-entropy} weight to be $1000$ compared to $1$ for non-argumentative sentences. We chose these weights based on label distribution in our training set described in Figure \ref{role_count}. Our best model is chosen based on the F1 score on the validation set.

\section{Effect of argument roles on generated summaries}
\label{sec:appendix}

Table \ref{example} shows an example of generated summaries with adding special tokens and without the special tokens. 

\begin{table*}[ht]
\begin{tabular}{|p{0.31\textwidth}|p{0.31\textwidth}|p{0.31\textwidth}|}
\hline
\textbf{Reference summary}                                                                                                                                                                                                                                                                                                                                                                                                                                                                                                                                                                                                                                                                                                                                                                                                                                                                                                                                                                                                                                                                                                                                                                                                                                                                                                                                                                                                                                                                                                                                                                                                                                                                                                                                                                                                                                                                                                                                                                                                                                                                                                                                    & \textbf{Vanilla LED-base}                                                                                                                                                                                                                                                                                                                                                                                                                                                                                                                                                                                                                                                                                                                                                                                                                                                                                                                                                                                                                                                                                                                                                                                                                                                                                                                                                                                                                                                                                                                                                                                                                                                                                                                                                                                                                                                                                                                                                                                 & \textbf{arg-LED-base "6 markers"}                                                                                                                                                                                                                                                                                                                                                                                                                                                                                                                                                                                                                                                                                                                                                                                                                                                          \\ \hline

The plaintiffs claim for crop damage caused by the defendant's cattle entering the plaintiff's canola field.  The defendant denies he is responsible or negligent, but admits his cattle were in the field. HELD: Damages were awarded in the amount of \$2,533.45. The court valued the loss at \$3,052.36. It then deducted 2\% attributed to wild animal damage, 5\% for the plaintiffs failure to calculate actual yield from the rest of the crop, 5\% that the plaintiffs would have paid in dockage and 5\% for the cost of production.  Civil liability for crop damage caused by cattle flows from the Stray Animals Act. A cattle owner is strictly liable for damages caused by his straying cattle.  This liability may possibly even be absolute and only an act of god may serve as a defence.  The issue here is the quantum of the damage. The fact that the defendant was refused access to the plaintiff's property to repair the fence has no bearing on liability in this case. An independent adjuster assessed the damage at 557 bushels. The court found on the facts that the damage was caused exclusively by the defendant's cattle on several occasions.                                                                                                                                                                                                                                                                                                                                                                                                                                                                                                                                                                                                                                                                                                                                                                                                                                                                                                      & The plaintiff's claim against the defendant is for crop damage sustained as a result of the defendant's cattle getting into the plaintiff's canola field in late July 1999. The plaintiff admits that his cattle were in the Plaintiff's crop on one occasion, but says that there is no evidence that their cattle are in the canola on any other occasion, and that such loss as was caused by his cattle is minimal. HELD: Judgment in the amount of \$2,533.45 was granted in favour of the plaintiffs, together with pre-judgment interest from January 2000 and costs. The present s.27 (1) has been in the Act since 1977, when the Act was substantially re-written, apparently in response to the increasing urbanization of the province. Under both the present and pre-1977 statutes, the legal situation was the same -- the cattle owner is strictly liable for damage caused by the straying cattle. The fact remains that the cattle did move onto the field unbeknownst to the defendant, his spouse, his brother, or his hired help. The Carrot River did not impede the cattle from moving into the field on more than one occasion and other cattle were not in the field.                                                                                                                                                                                                                                                                                                                                                                                                                                                                                                                                                                                                                                                                                                                                                                                                     &  The plaintiff's claim against the defendant was for crop damage sustained as a result of his cattle getting into the plaintiff's canola field in late July 1999.  The defendant admits that his cattle were in the Plaintiff's crop on one occasion, but says that there is no evidence that any other cattle were there.  HELD: Judgment in the amount of \$2,533.45 was granted in favour of the plaintiffs, together with pre-judgment interest from January 2000 and costs. Under both the present s.27 (1) and the pre-1977 statutes, the legal situation was the same -- the cattle owner is strictly liable for damage caused by his straying cattle.  \\ \hline

\end{tabular}
\caption{\label{example}Example of generated summaries with \textbf{Vanilla LED-base} and \textbf{arg-LED-base} versus reference summary.}
\end{table*}

\end{document}